# Identifying the Class of Maxi-Consistent Operators in Argumentation

**Srdjan Vesic**　　　　　　　　　　　　　　　　　　　　　　　　　　　　　VESIC@CRIL.FR
*CRIL - CNRS*
*Rue Jean Souvraz SP 18*
*F 62307 Lens Cedex*
*FRANCE*

## Abstract

Dung's abstract argumentation theory can be seen as a general framework for non-monotonic reasoning. An important question is then: what is the class of logics that can be subsumed as instantiations of this theory? The goal of this paper is to identify and study the large class of logic-based instantiations of Dung's theory which correspond to the maxi-consistent operator, i.e. to the function which returns maximal consistent subsets of an inconsistent knowledge base. In other words, we study the class of instantiations where every extension of the argumentation system corresponds to exactly one maximal consistent subset of the knowledge base. We show that an attack relation belonging to this class must be conflict-dependent, must not be valid, must not be conflict-complete, must not be symmetric etc. Then, we show that some attack relations serve as lower or upper bounds of the class (e.g. if an attack relation contains *canonical undercut* then it is not a member of this class). By using our results, we show for all existing attack relations whether or not they belong to this class. We also define *new* attack relations which are members of this class. Finally, we interpret our results and discuss more general questions, like: what is the added value of argumentation in such a setting? We believe that this work is a first step towards achieving our long-term goal, which is to better understand the role of argumentation and, particularly, the expressivity of logic-based instantiations of Dung-style argumentation frameworks.

## 1. Introduction

A question whether Dung's (1995) abstract theory can be used as a general framework for non-monotonic reasoning has drawn a particular amount of attention among researchers in artificial intelligence. More precisely, the question is: can existing or new approaches to reasoning be seen as instantiations of Dung's theory? This is certainly a very general question. Furthermore, different approaches suppose that the available knowledge is represented in different form. This paper studies the problem setting when one is given a finite inconsistent set of classical propositional logic formulae, which we refer to as a *knowledge base*. There are a number of approaches for dealing with inconsistent information: a notable example are paraconsistent logics (Priest, 2002) where one is able to draw some (but not all) conclusions from an inconsistent set of formulae. Indeed, each paraconsistent logic allows for a subset of the inferences that could be obtained using classical logic with the same knowledge. Other examples of dealing with inconsistent knowledge include belief revision (Gardenfors, 1988), belief merging (Konieczny & Pérez, 2011) or voting (Arrow, Sen, & Suzumura, 2002). To be completely precise, note that there are approaches where one is given a *multiset* instead of a set, for example where several voters can express their knowledge or preferences and the number of agents stating / voting for a proposition is important. However, in this paper, we suppose that the information is represented in form of a set.

Generally speaking, we call an *operator* a function which provides a way to go from an inconsistent knowledge base to a set of subsets of that knowledge base. Examples of operators are: a function returning maximal for set inclusion consistent subsets of a knowledge base, called *maxi-consistent operator*, a function returning maximal for cardinality consistent subsets of a knowledge base, called *maxi-card operator*, a function returning *all* consistent subsets of a knowledge base...

To understand how and to which extent Dung's theory can be used as a general framework for reasoning, it is essential to study the link between the result obtained by applying an operator to a knowledge base $\Sigma$ and the extensions of the argumentation framework $\mathcal{F} = (\texttt{Arg}(\Sigma), \mathcal{R})$, where for a set $S \subseteq \Sigma$, we denote by $\texttt{Arg}(S)$ the set of all arguments that can be built from $S$, and by $\mathcal{R}$ the relation used for identifying attacks between arguments. There are papers (Cayrol, 1995; Caminada & Amgoud, 2007; Amgoud & Besnard, 2009, 2010; Amgoud & Vesic, 2010; Gorogiannis & Hunter, 2011) studying the notions which are somehow related to a link between a knowledge base and the corresponding argumentation framework. However, since the work of Dung (1995), there are almost no papers studying the link between an *operator* and an *instantiation* of Dung's theory. Cayrol (1995) showed that the instantiation of Dung's theory using *stable semantics* and *direct undercut* as attack relation, corresponds to maxi-consistent operator. In argumentation community, this one-to-one correspondence is sometimes identified as a main objection against pure logic-based argumentation, because the additional value of constructing the argumentation framework is then said to be questionable (since computing the extensions does not do more than applying maxi-consistent operator). However, a recent work by Vesic and van der Torre (2012) shows that there exists a large class of logic-based instantiations of Dung's abstract theory having two interesting features: (i) it returns extensions that do *not* correspond to maximal consistent subsets of the initial knowledge base, and (ii) its result satisfies basic argumentation postulates (Caminada & Amgoud, 2007), e.g. consistency, closure... That paper shows that the space of logic-based instantiations of Dung's theory is much larger than it was believed.

The previous result makes the question "what is the class of operators that can be viewed as instantiations of Dung's theory?" even more relevant as a research topic. We now aim only at giving a broad overview of this class. First note that, interestingly, there is a rather big class of instantiations of Dung's theory returning inconsistent results, as showed by Caminada and Amgoud (2007). However, one would normally prefer to avoid this type of behaviour, and to study the class of instantiations returning consistent results.[1] Thus, our long-term goal is to identify the whole class of instantiations of Dung's theory that yield a consistent result. However, that is certainly a hard task. We start by noticing that, given a set $\Sigma$, the most common and a well-known way to deal with inconsistent information is to use the maxi-consistent operator, i.e. to select maximal consistent subsets of $\Sigma$. Also, we conjecture that one of the "biggest"[2] sub-classes of the class of instantiations returning a consistent result is the class of instantiations corresponding to maxi-consistent operator. That is why our first goal, and the main goal of this paper, is to study this class. (Note that a more general approach would consider the set of maxi-consistent subsets of $\Sigma$ *and* a selection function $f$ among the maxi-consistent subsets. Thus, only some maxi-consistent sets would be used for reasoning. However, the present paper studies the case when all the maxi-consistent subsets of $\Sigma$ are taken into account since this already captures a significant number of systems.)

---

1. Note that while there is no consensus regarding some of the postulates (e.g. indirect consistency), some postulates (e.g. direct consistency) enjoy much wider acceptance.
2. informally, but in the sense: in number of known instantiations

More particularly, the paper aims to answering the following questions: What are the properties of attack relations and semantics used in instantiations of Dung's theory that correspond to maxi-consistent operators? What are the necessary and sufficient conditions so that an attack relation belongs to this class (under a given semantics)? What properties do they satisfy: must (not) they be conflict-dependent, valid, symmetric, ... ? Can we identify sub-classes of attack relations belonging / not belonging to this class? Can we find the lower and / or the upper bound (in terms of set inclusion) of this class of attack relations? Which existing (in the literature) attack relations belong to this class under which semantics? Are there new attack relations belonging to this class?

The paper is organised as follows: Section 2 introduces the main notions of argumentation theory we use in the rest of the paper. Section 3 formally defines the class of instantiations corresponding to maxi-consistent operator. Section 4 shows what properties are satisfied by attack relations belonging to this class. Section 5 identifies several classes of attack relations (not-)corresponding to maxi-consistent operator. Section 6 shows for all (to the best of our knowledge) existing attack relations from literature whether or not they belong to this class, and also defines a new attack relation which is a member of this class. The last section concludes and discusses related work.

## 2. Basics of Argumentation

As already mentioned, this paper supposes that one is given a set of classical propositional logic formulae $\Sigma$. We use the well-known (e.g. Besnard & Hunter, 2001; Amgoud & Cayrol, 2002; Gorogiannis & Hunter, 2011) *logic-based* approach for instantiating Dung's theory. $\mathcal{L}$ denotes the set of well-formed formulae, $\vdash$ stands for classical entailment, and $\equiv$ for logical equivalence. We use the notation $\text{MC}(\Sigma)$ for the set of all maximal consistent subsets of $\Sigma$.

A logical argument is defined as a pair $(support, conclusion)$.

**Definition 1** (Argument). *An argument is a pair $(\Phi, \alpha)$ such that $\Phi \subseteq \Sigma$ is a minimal (for set inclusion) consistent set of formulae such that $\Phi \vdash \alpha$.*

For an argument $a = (\Phi, \alpha)$, we use the function $\text{Supp}(a) = \Phi$ to denote its support and $\text{Conc}(a) = \alpha$ to denote its conclusion.

**Example 1.** *Let $\Sigma = \{\varphi, \varphi \to \psi, \omega\}$. $a = (\{\varphi, \varphi \to \psi\}, \psi)$, $b = (\{\varphi \to \psi\}, \neg\varphi \vee \psi)$ and $c = (\{\varphi, \psi\}, \varphi \leftrightarrow \psi)$ are some of the arguments that can be constructed from $\Sigma$. For example, $\text{Supp}(a) = \{\varphi, \varphi \to \psi\}$ and $\text{Conc}(a) = \psi$.*

For a given set of formulae $S$, we denote by $\text{Arg}(S)$ the set of arguments constructed from $S$. Formally, $\text{Arg}(S) = \{a \mid a \text{ is an argument and } \text{Supp}(a) \subseteq S\}$. Let $\text{Arg}(\mathcal{L})$ denote the set of all arguments that can be constructed from the language of propositional logic. For a given set of arguments $\mathcal{E}$, we denote $\text{Base}(\mathcal{E}) = \bigcup_{a \in \mathcal{E}} \text{Supp}(a)$. We suppose that function $\text{Arg}$ is defined on $\mathcal{L}$ and that function $\text{Base}$ is defined on $\text{Arg}(\mathcal{L})$; by slightly abusing the notation, we sometimes write $\text{Arg}$ (respectively $\text{Base}$) for the restriction of these functions on any set of formulae (respectively arguments).

**Definition 2** (Argumentation system). *An argumentation system (AS) is a pair $(\mathcal{A}, \mathcal{R})$ where $\mathcal{A} \subseteq \text{Arg}(\mathcal{L})$ is a set of arguments and $\mathcal{R} \subseteq \mathcal{A} \times \mathcal{A}$ a binary relation. For each pair $(a, b) \in \mathcal{R}$, we say that $a$ attacks $b$. We also sometimes use notation $a\mathcal{R}b$ instead of $(a, b) \in \mathcal{R}$.*

In order to simplify notation, we do not explicitly mention an argumentation system when it is clear from the context which argumentation system we refer to. Since arguments are built from formulae, we suppose that an attack relation is defined by specifying a condition such that for every two arguments $a$ and $b$, we have that $a$ attacks $b$ if and only if the condition from the definition of attack relation is satisfied. For example, such a condition can be that the conclusion of $a$ is logically equivalent to the negation of the conclusion of $b$. We suppose that all attack relations are defined on the set $\text{Arg}(\mathcal{L}) \times \text{Arg}(\mathcal{L})$, and that for every set $\mathcal{A} \subseteq \text{Arg}(\mathcal{L})$, we use the restriction of the attack relation on the set $\mathcal{A} \times \mathcal{A}$. That is why, in order to simplify notation, we simply write $\mathcal{R}$ for an attack relation defined on the set $\text{Arg}(\mathcal{L}) \times \text{Arg}(\mathcal{L})$ as well as for the restriction of that attack relation on every set $\mathcal{A} \times \mathcal{A}$, with $\mathcal{A} \subseteq \text{Arg}(\mathcal{L})$.

In order to determine mutually acceptable sets of arguments, different semantics have been introduced in argumentation. We first introduce the basic notions of conflict-freeness and defence.

**Definition 3** (Conflict-free, defence). *Let $\mathcal{F} = (\mathcal{A}, \mathcal{R})$ be an AS, $\mathcal{E} \subseteq \mathcal{A}$ and $a \in \mathcal{A}$.*

- *$\mathcal{E}$ is* conflict-free *if and only if there exist no arguments $a, b \in \mathcal{E}$ such that $a \, \mathcal{R} \, b$*

- *$\mathcal{E}$ defends $a$ if and only if for every $b \in \mathcal{A}$ we have that if $b \, \mathcal{R} \, a$ then there exists $c \in \mathcal{E}$ such that $c \, \mathcal{R} \, b$.*

Let us now define the most commonly used semantics.

**Definition 4** (Acceptability semantics). *Let $\mathcal{F} = (\mathcal{A}, \mathcal{R})$ be an AS and $\mathcal{B} \subseteq \mathcal{A}$. We say that a set $\mathcal{B}$ is* admissible *if and only if it is conflict-free and defends all its elements.*

- *$\mathcal{B}$ is a* complete *extension if and only if $\mathcal{B}$ defends all its arguments and contains all the arguments it defends.*

- *$\mathcal{B}$ is a* preferred *extension if and only if it is a maximal (with respect to set inclusion) admissible set.*

- *$\mathcal{B}$ is a* stable *extension if and only if $\mathcal{B}$ is conflict-free and for all $a \in \mathcal{A} \setminus \mathcal{B}$, there exists $b \in \mathcal{B}$ such that $b \, \mathcal{R} \, a$.*

- *$\mathcal{B}$ is a* semi-stable *extension if and only if $\mathcal{B}$ is a complete extension and the union of the set $\mathcal{B}$ and the set of all arguments attacked by $\mathcal{B}$ is maximal (for set inclusion).*

- *$\mathcal{B}$ is a* grounded *extension if and only if $\mathcal{B}$ is a minimal (for set inclusion) complete extension.*

- *$\mathcal{B}$ is an* ideal *extension if and only if $\mathcal{B}$ is a maximal (for set inclusion) admissible set contained in every preferred extension.*

For an argumentation system $\mathcal{F} = (\mathcal{A}, \mathcal{R})$ we denote by $\text{Ext}_x(\mathcal{F})$; or, by a slight abuse of notation, by $\text{Ext}_x(\mathcal{A}, \mathcal{R})$ the set of its extensions with respect to semantics $x$. We use abbreviations $c$, $p$, $s$, $ss$, $g$ and $i$ for respectively complete, preferred, stable, semi-stable, grounded and ideal semantics. For example, $\text{Ext}_p(\mathcal{F})$ denotes the set of preferred extensions argumentation system $\mathcal{F}$.

**Example 2.** *Let $\mathcal{F} = (\mathcal{A}, \mathcal{R})$ be an argumentation framework with $\mathcal{A} = \{a, b, c, d\}$ and $\mathcal{R} = \{(b, c), (c, b), (b, d), (c, d)\}$. There are two preferred/stable/semi-stable extensions: $\{a, b\}$ and $\{a, c\}$; three complete extensions: $\{a\}$, $\{a, b\}$ and $\{a, c\}$; and one grounded/ideal extension: $\{a\}$.*

## 3. Defining the Problem Setting

Until now, we specified how to, from a knowledge base $\Sigma$, construct an argumentation system $\mathcal{F} = (\texttt{Arg}(\Sigma), \mathcal{R})$, and then, using a chosen semantics, calculate extensions. Since all the components of the system except a semantics and an attack relation are fixed, then whether an instantiation corresponds to maxi-consistent operator depends exclusively on those two components. The next definition provides a formal definition of what we mean by saying that an instantiation of Dung's framework "corresponds" to maxi-consistent operator. The idea is that the function $\texttt{Arg}$ should be a bijection between $\texttt{MC}(\Sigma)$ and the extensions of the corresponding argumentation system.

**Definition 5** (**MC $\leftrightarrow$ Ext**). *Let $x$ be an argumentation semantics. We say that attack relation $\mathcal{R}$ satisfies* ($\texttt{MC} \leftrightarrow \texttt{Ext}_x$) *if and only if for every finite set of propositional formulae $\Sigma$ we have that*

$$\texttt{Arg is a bijection between } \texttt{MC}(\Sigma) \text{ and } \texttt{Ext}_x(\texttt{Arg}(\Sigma), \mathcal{R})$$

This means that for every set $S \in \texttt{MC}(\Sigma)$, it holds that $\texttt{Arg}(S) \in \texttt{Ext}(\texttt{Arg}(\Sigma), \mathcal{R})$ and for every $\mathcal{E} \in \texttt{Ext}(\texttt{Arg}(\Sigma), \mathcal{R})$, there exists $S \in \texttt{MC}(\Sigma)$ such that $\mathcal{E} = \texttt{Arg}(S)$. For example, we say that an attack relation $\mathcal{R}$ satisfies ($\texttt{MC} \leftrightarrow \texttt{Ext}_c$) if and only if for every finite $\Sigma$, we have that $\texttt{Arg}$ is a bijection between $\texttt{MC}(\Sigma)$ and $\texttt{Ext}_c(\texttt{Arg}(\Sigma), \mathcal{R})$. Sometimes, when it is clear from the context which semantics we refer to or when a semantics is not important, we use the simplified notation ($\texttt{MC} \leftrightarrow \texttt{Ext}$). We say that an attack relation $\mathcal{R}$ falsifies ($\texttt{MC} \leftrightarrow \texttt{Ext}_x$) if and only if $\mathcal{R}$ does not satisfy ($\texttt{MC} \leftrightarrow \texttt{Ext}_x$). The following example shows an attack relation that does not satisfy ($\texttt{MC} \leftrightarrow \texttt{Ext}_s$).

**Example 3.** *Consider the attack relation known as* defeating rebut, *denoted by $\mathcal{R}_{dr}$ and defined as follows: for two arguments $a$ and $b$, we say that $a$ attacks $b$ and write $a\mathcal{R}_{dr}b$ if and only if $\texttt{Conc}(a) \vdash \neg\texttt{Conc}(b)$. This attack relation falsifies* ($\texttt{MC} \leftrightarrow \texttt{Ext}_s$). *To see why, it is sufficient to find a set of formulae $\Sigma$ such that $\texttt{Arg}$ is not a bijection between $\texttt{MC}(\Sigma)$ and $(\texttt{Arg}(\Sigma), \mathcal{R}_{dr})$. To that end, consider $\Sigma = \{\varphi \wedge \psi, \varphi \wedge \neg\psi\}$ and denote $\mathcal{F} = (\texttt{Arg}(\Sigma), \mathcal{R}_{dr})$. We see that $\texttt{MC}(\Sigma) = \{S_1, S_2\}$ with $S_1 = \{\varphi \wedge \psi\}$ and $S_2 = \{\varphi \wedge \neg\psi\}$. Denote $\mathcal{E}_1 = \texttt{Arg}(S_1)$ and $\mathcal{E}_2 = \texttt{Arg}(S_2)$. If $\texttt{Ext}_s(\mathcal{F}) \neq \{\mathcal{E}_1, \mathcal{E}_2\}$ then $\mathcal{R}_{dr}$ does not satisfy* ($\texttt{MC} \leftrightarrow \texttt{Ext}_s$). *Consider argument $a = (\{\varphi \wedge \psi\}, \varphi \wedge \psi)$, and note that $a \in \mathcal{E}_1$. Observe that for every argument $b \in \texttt{Arg}(\Sigma)$, we have that $b\mathcal{R}_{dr}a$ if and only if $\texttt{Conc}(b) \vdash \neg(\varphi \vee \psi)$. In other words, for every argument $b \in \texttt{Arg}(\Sigma)$, $b$ attacks $a$ if and only if $\texttt{Conc}(b) \vdash \neg\varphi \wedge \neg\psi$. Recall that from Definition 1 we know that for every argument $b$, $\texttt{Supp}(b) \vdash \texttt{Conc}(b)$. Thus, for every argument $b \in \texttt{Arg}(\Sigma)$, if $b\mathcal{R}_{dr}a$ then $\texttt{Supp}(b) \vdash \neg\varphi \wedge \neg\psi$. Since $\Sigma = \{\varphi \wedge \psi, \varphi \wedge \neg\psi\}$ then there is no argument $b \in \texttt{Arg}(\Sigma)$ such that $\texttt{Supp}(b) \vdash \neg\varphi \wedge \neg\psi$. Thus, argument $a$ is not attacked by any argument of $\mathcal{E}_2$. This means that $\mathcal{E}_2$ is not a stable extension of $\mathcal{F}$. Consequently, $\texttt{Arg}$ is not a bijection between $\texttt{MC}(\Sigma)$ and $\texttt{Ext}_s(\mathcal{F})$. Hence, $\mathcal{R}_{dr}$ falsifies* ($\texttt{MC} \leftrightarrow \texttt{Ext}_s$).

### 3.1 Complete and Incomplete Systems

There are two ways to study the link between an instantiated argumentation system $\mathcal{F}$ (containing arguments and attacks between them) and the corresponding knowledge base $\Sigma$ (containing formulae). The *first scenario* is as follows:

- choose an attack relation $\mathcal{R}$ and a semantics $x$
- start with a finite knowledge base $\Sigma$

- consider the system $\mathcal{F} = (\text{Arg}(\Sigma), \mathcal{R})$, containing all the arguments that can be built from $\Sigma$

- compare the result obtained by using an operator on $\Sigma$ and the one obtained by calculating the extensions of $\mathcal{F}$

In this case, we say that the obtained argumentation system is *complete*. Every complete system has an infinite number of arguments, but for every complete system $\mathcal{F}$, there exists a finite system $\mathcal{F}'$ such that $\mathcal{F}$ and $\mathcal{F}'$ are equivalent. How to formally define equivalence between argumentation systems is not the topic of the present paper; for more details the reader is invited to consult the literature on this subject (Amgoud, Besnard, & Vesic, 2011).

The *second possibility* is to do the converse:

- for a given attack relation $\mathcal{R}$ and semantics $x$,

- start with an argumentation system $\mathcal{F} = (\mathcal{A}, \mathcal{R})$,

- define $\Sigma$ as the set of all formulae used in the supports of arguments of $\mathcal{F}$, that is, define $\Sigma \stackrel{\text{def}}{=} \text{Base}(\mathcal{A})$

- compare the result obtained by using an operator on $\Sigma$ and the one obtained by calculating the extensions of $\mathcal{F}$

The obtained argumentation system may be *incomplete*, in the sense that $\mathcal{A} \neq \text{Arg}(\text{Base}(\mathcal{A}))$. There is an important difference between those two scenarios. Namely, in the first case, all the arguments that can be built from $\Sigma$ are considered when calculating $\text{Ext}_x(\mathcal{F})$. In the second case, $\Sigma$ contains all the formulae from $\mathcal{A}$, but in $\mathcal{F}$, not all formulae are equally represented. Let us illustrate this situation.

**Example 4.** *Let $\mathcal{R}$ be defined as: for every $a, b \in \text{Arg}(\mathcal{L})$, $(a, b) \in \mathcal{R}$ if and only if there exists $\varphi \in \text{Supp}(b)$ such that $\text{Conc}(a) \equiv \neg\varphi$. Let us use preferred semantics. Let $\mathcal{F} = (\mathcal{A}, \mathcal{R})$ and $\mathcal{A} = \{a, b\}$ with $a = (\{\varphi, \varphi \to \psi\}, \psi)$ and $b = (\{\neg\varphi\}, \neg\varphi)$. In this case, since $b$ attacks $a$ and not vice versa, the only extension is $\mathcal{E} = \{b\}$. Note that the conclusion of the only accepted argument is $\neg\varphi$. However, if we take $\Sigma$ to be the union of all formulae used in supports of the arguments of $\mathcal{F}$, we obtain $\Sigma = \{\varphi, \neg\varphi, \varphi \to \psi\}$. There are two maximal consistent subsets of this knowledge base: $\text{MC}(\Sigma) = \{\{\varphi, \varphi \to \psi\}, \{\neg\varphi, \varphi \to \psi\}\}$.*

It is clear that in a setting similar to that in the previous example, or, more generally, in the second scenario, one cannot expect Arg to be a bijection between $\text{MC}(\Sigma)$ and $\text{Ext}(\mathcal{A}, \mathcal{R})$. But, what does an incomplete argumentation system stand for? How is it obtained? To conclude that a system such that $\mathcal{A} \neq \text{Arg}(\text{Base}(\mathcal{A}))$ is meaningless would certainly be too hasty. Let us consider this question in more detail. Namely, we know that "missing" arguments can be added by an intelligent agent. Should we first add all the "missing" arguments and then calculate the extensions of the complete version of our system? There are two possible answers: (1) yes, we must add missing arguments in order to take into account all available information; (2) no, since we are given an argumentation system and not all arguments have been constructed (in case of monological argumentation) or uttered (in the case of dialogical argumentation). Both arguments (1) and (2) make sense in different applications: the first possibility corresponds to a case when we want to simulate a resource unbounded agent, and take into account all the information (where "information" is seen as formulae) known by agent(s). Note that this has its disadvantages since by doing so, we ignore the argumentational representation of the problem. The second possibility is to be used when we want to know what the output of an argumentation system is, where we do take into account the fact that not all arguments are constructed (e.g. because of the lack of computational resources, or since the given argumentation framework is representing a dialogue in which not everything has been said).

Note that numerous works in the 1990s (Pollock, 1992; Vreeswijk, 1997; Loui, 1998) yield conceptual and philosophical arguments supporting partial computation (i.e. incomplete systems). An important part of Vreeswijk's (1997) work is devoted to defining and constructing complete argumentation systems. Loui (1998) discusses the philosophical difference between demonstrative reasoning and non-demonstrative reasoning and claims that in a realistic (i.e. resource-bounded) setting, not all reasons are demonstrative, and that *process* and *disputation* are essential to reasoning. Note, however, that none of those frameworks is an instantiation of Dung's system, and that formalisations in those works differ a lot from the framework we studied in this paper. The goal of the present paper is not to argue that complete systems are in any sense "better" than incomplete ones (or vice versa), but only to study the possibilities and limits related to instantiating Dung's abstract theory. We will not further analyse the difference between complete and incomplete systems, but we find it necessary to point out that they exist, in order to make the context of our research question clear. In the second scenario, it is not reasonable to expect any correlation between the result obtained directly from $\Sigma$ and from $\mathcal{F}$. That is why, in the rest of the paper we suppose the first scenario.

## 4. Properties of Relations Satisfying (MC↔Ext)

In this section, we analyse properties of attack relations satisfying $(\text{MC} \leftrightarrow \text{Ext})$. We first show that if this condition is satisfied, then the function $\text{Base} : \text{Ext}(\mathcal{F}) \to \text{MC}(\Sigma)$ is the inverse function of the function $\text{Arg} : \text{MC}(\Sigma) \to \text{Ext}(\mathcal{F})$.

**Proposition 1.** *Let $\mathcal{R}$ be an attack relation and $x$ an acceptability semantics. If relation $\mathcal{R}$ satisfies $(\text{MC} \leftrightarrow \text{Ext}_x)$ then:*

- *for every $S \in \text{MC}(\Sigma)$, we have that $S = \text{Base}(\text{Arg}(S))$,*
- *for every $\mathcal{E} \in \text{Ext}_x(\mathcal{F})$, we have that $\mathcal{E} = \text{Arg}(\text{Base}(\mathcal{E}))$.*

*Proof.* Let $\Sigma$ be a finite set of propositional formulae and let $\mathcal{F} = (\text{Arg}(\Sigma), \mathcal{R})$.

- Let $S \in \text{MC}(\Sigma)$ and $\mathcal{E} = \text{Arg}(S)$. Since $\mathcal{R}$ satisfies $(\text{MC} \leftrightarrow \text{Ext})$, then $\mathcal{E} \in \text{Ext}_x(\mathcal{F})$. Let $S' = \text{Base}(\mathcal{E})$ and let us suppose that $S \neq S'$. Let us study two cases.
    - Let $S \setminus S' \neq \emptyset$ and $\alpha' \in S \setminus S'$. This means that there is an argument $a \in \mathcal{E}$ such that $\alpha' \in \text{Supp}(a)$ with $\alpha' \notin S$, contradiction.
    - Let $S' \setminus S \neq \emptyset$ and $\alpha' \in S' \setminus S$. This means that there is an argument $a \in \mathcal{E}$ such that $\alpha' \in \text{Supp}(a)$. Contradiction with $\alpha' \notin S$.

  Since $S \setminus S' = \emptyset$ and $S' \setminus S = \emptyset$, then $S = S'$; In other words, $\text{Base}(\text{Arg}(S)) = S$.

- Let $\mathcal{E} \in \text{Ext}_x(\mathcal{F})$ and $S = \text{Base}(\mathcal{E})$. Since $\mathcal{R}$ satisfies $(\text{MC} \leftrightarrow \text{Ext}_x)$, then there exists a unique $S' \in \text{MC}(\Sigma)$ such that $\text{Arg}(S') = \mathcal{E}$. Let us prove that $S = S'$.

- Let us suppose that $S \setminus S' \neq \emptyset$ and let $\alpha \in S \setminus S'$. This means that there is an argument $a \in \mathcal{E}$ such that $\alpha \in \text{Supp}(a)$. Contradiction with the fact $\alpha \notin S'$.

- Suppose that $S' \setminus S \neq \emptyset$ and that $\alpha' \in S' \setminus S$. From $\alpha' \in S'$, we conclude that there exists $a \in \mathcal{E}$ such that $\alpha' \in \text{Supp}(a)$. Contradiction with the fact that $\alpha' \notin S$.

From $S \setminus S' = \emptyset$ and $S' \setminus S = \emptyset$, we conclude $S = S'$. Thus, $\text{Arg}(\text{Base}(\mathcal{E})) = \mathcal{E}$.

□

Let us illustrate this result by the following example.

**Example 5.** *Consider the attack relation known as* direct undercut, *denoted by* $\mathcal{R}_{du}$ *and defined as follows: for two arguments $a$ and $b$, we say that $a$ attacks $b$ and we write $a\mathcal{R}_{du}b$ if and only if there exists $\varphi \in \text{Supp}(b)$ such that $\text{Conc}(a) \equiv \neg\varphi$. It is known that* direct undercut *satisfies* $(\text{MC} \leftrightarrow \text{Ext}_s)$ *(Cayrol, 1995). From Proposition 1, we see that for every $\Sigma$, for every $S \in \text{MC}(\Sigma)$, it holds that $S = \text{Base}(\text{Arg}(S))$ and, more interestingly, that for every $\mathcal{E} \in \text{Ext}_s(\text{Arg}(\Sigma), \mathcal{R}_{du})$, we have that $\mathcal{E} = \text{Arg}(\text{Base}(\mathcal{E}))$.*

The previous result allows to easily show that if an attack relation satisfies $(\text{MC} \leftrightarrow \text{Ext})$, then every extension has a consistent base and the union of its arguments' conclusions is consistent.

**Corollary 1.** *Let $\mathcal{R}$ be an attack relation and $x$ a semantics. Let $\mathcal{R}$ satisfy $(\text{MC} \leftrightarrow \text{Ext}_x)$ and let $\Sigma$ be a finite set of formulae. Denote $\mathcal{F} = (\text{Arg}(\Sigma), \mathcal{R})$. Then, for every $\mathcal{E} \in \text{Ext}_x(\mathcal{F})$, we have:*

- $\text{Base}(\mathcal{E})$ *is consistent*
- $\bigcup_{a \in \mathcal{E}} \text{Conc}(a)$ *is consistent*

*Proof.* Let $\mathcal{E} \in \text{Ext}_x(\mathcal{F})$. Since $\mathcal{R}$ satisfies $(\text{MC} \leftrightarrow \text{Ext}_x)$, then there exists $S \in \text{MC}(\Sigma)$ such that $\mathcal{E} = \text{Arg}(S)$. From Proposition 1, we obtain $\mathcal{E} = \text{Arg}(\text{Base}(\mathcal{E}))$. Since $\text{Arg}$ is an injective function, for every $S' \in \text{MC}(\Sigma)$, if $\mathcal{E} = \text{Arg}(S')$ then $S = S'$. Thus, $S = \text{Base}(\mathcal{E})$. Consequently, $\text{Base}(\mathcal{E})$ is a consistent set. It is clear that for every argument $a \in \mathcal{E}$, we have that $\text{Base}(\mathcal{E}) \vdash \text{Conc}(a)$. Since $\text{Base}(\mathcal{E})$ is consistent, then $\bigcup_{a \in \mathcal{E}} \text{Conc}(a)$ is consistent as well. □

Note that we can use the previous result to show that an attack relation does *not* satisfy $(\text{MC} \leftrightarrow \text{Ext})$. Namely, if an attack relation returns extensions having inconsistent bases, then it violates $(\text{MC} \leftrightarrow \text{Ext})$.

**Corollary 2.** *Let $\mathcal{R}$ be an attack relation, and $x$ an acceptability semantics. If there exists a finite knowledge base $\Sigma$ such that there exists an extension $\mathcal{E} \in \text{Ext}_x(\text{Arg}(\Sigma), \mathcal{R})$ such that $\text{Base}(\mathcal{E})$ is inconsistent, then $\mathcal{R}$ does not satisfy $(\text{MC} \leftrightarrow \text{Ext}_x)$.*

### 4.1 On Conflict-Dependence and Validity

In this subsection, we study the link between satisfying $(\text{MC} \leftrightarrow \text{Ext})$ and conflict-dependence and validity. An attack relation is *conflict-dependent* if whenever an argument attacks another one, the union of their supports is inconsistent (Amgoud & Besnard, 2009).

**Definition 6** (Conflict-dependent). *Let $\mathcal{R} \subseteq \text{Arg}(\mathcal{L}) \times \text{Arg}(\mathcal{L})$ be an attack relation. We say that $\mathcal{R}$ is conflict-dependent if and only if for every $a, b \in \text{Arg}(\mathcal{L})$, if $(a, b) \in \mathcal{R}$ then $\text{Supp}(a) \cup \text{Supp}(b) \vdash \bot$.*

We now prove that conflict-dependence is a necessary condition for satisfying (MC ↔ Ext). To be completely precise, we here specify that we say that a *semantics $x$ returns conflict-free sets* if and only if for every argumentation system $(\mathcal{A}, \mathcal{R})$, for every $\mathcal{E} \in \text{Ext}_x(\mathcal{A}, \mathcal{R})$, it holds that $\mathcal{E}$ is conflict-free with respect to $\mathcal{R}$. All the semantics from Definition 4 return conflict-free sets.

**Proposition 2.** *Let $\mathcal{R}$ be an attack relation and $x$ a semantics returning conflict-free sets. If $\mathcal{R}$ satisfies (MC ↔ $\text{Ext}_x$), then $\mathcal{R}$ is conflict-dependent.*

*Proof.* Let us suppose the contrary, i.e. let $\mathcal{R}$ be an attack relation that is not conflict-dependent, let $\Sigma$ be a knowledge base and let $a, b \in \text{Arg}(\Sigma)$ with $a\mathcal{R}b$ and $\text{Supp}(a) \cup \text{Supp}(b)$ being consistent. Thus, there exists a set $S \in \text{MC}(\Sigma)$ such that $\text{Supp}(a) \cup \text{Supp}(b) \subseteq S$. Since $\mathcal{R}$ satisfies (MC ↔ $\text{Ext}_x$) then $\mathcal{E} = \text{Arg}(S)$ is an extension of the corresponding argumentation system $\mathcal{F} = (\text{Arg}(\Sigma), \mathcal{R})$. This means that $a, b \in \mathcal{E}$. Contradiction with the assumption that $x$ returns conflict-free extensions. Thus, $\mathcal{R}$ must be conflict-dependent. □

Having proved this, we know that a relation satisfying (MC ↔ Ext) enjoys all the properties of conflict-dependent relations. For example, it was shown that if an attack relation is conflict-dependent, then there are no self-attacking arguments (Amgoud & Besnard, 2009).

**Corollary 3.** *Let $\mathcal{R}$ be an attack relation and $x$ a semantics returning conflict-free sets. If $\mathcal{R}$ satisfies (MC ↔ $\text{Ext}_x$) then for every argument $a \in \text{Arg}(\mathcal{L})$, we have that such that $(a, a) \notin \mathcal{R}$.*

*Proof.* From Proposition 2, we have that $\mathcal{R}$ is conflict-dependent. Then, there are no self-attacking arguments (Amgoud & Besnard, 2009, Prop. 4). □

This means that we have another way to identify (some of the) attack relations not satisfying (MC ↔ Ext): namely, if for an attack relation there exists a self-attacking argument, then the given attack relation falsifies (MC ↔ Ext) for all semantics returning conflict-free sets. Let us now study the notion of validity (Amgoud & Besnard, 2010).

**Definition 7** (Valid). *Let $\mathcal{R} \subseteq \text{Arg}(\mathcal{L}) \times \text{Arg}(\mathcal{L})$ be an attack relation. We say that $\mathcal{R}$ is valid if and only if for every $\mathcal{E} \subseteq \text{Arg}(\mathcal{L})$ it holds that if $\mathcal{E}$ is conflict-free, then $\text{Base}(\mathcal{E})$ is consistent.*

Let us now show that this property is incompatible with conflict-dependence.

**Proposition 3.** *There exists no attack relation which is both conflict-dependent and valid.*

*Proof.* Let $\mathcal{R}$ be an attack relation and suppose that $\mathcal{R}$ is both conflict-dependent and valid. Let $a = (\{\varphi\}, \varphi)$, $b = (\{\neg\psi\}, \neg\psi)$, $c = (\{\varphi \to \psi\}, \neg\varphi \lor \psi)$ and let $\mathcal{E} = \{a, b, c\}$. Since $\mathcal{R}$ is valid and $\text{Base}(\mathcal{E})$ is inconsistent, then $\mathcal{E}$ is not conflict-free. Since $\mathcal{R}$ is conflict-dependent, then $(a, b) \notin \mathcal{R}, (b, a) \notin \mathcal{R}, (a, c) \notin \mathcal{R}, (c, a) \notin \mathcal{R}, (b, c) \notin \mathcal{R}, (c, b) \notin \mathcal{R}$. Thus, $\mathcal{E}$ is conflict-free. Contradiction. □

This means that if an attack relation $\mathcal{R}$ satisfies (MC ↔ Ext) then there must exist a set $\mathcal{E}$ which is conflict-free with respect to $\mathcal{R}$ but whose base is inconsistent.

**Corollary 4.** *Let $\mathcal{R}$ be an attack relation and $x$ an acceptability semantics returning conflict-free sets and let $\mathcal{R}$ satisfy (MC ↔ $\text{Ext}_x$). Then, $\mathcal{R}$ is not valid.*

The proof of the previous fact is a consequence of Proposition 2 and Proposition 3. It is useful since if an attack relation is valid, we can immediately conclude that it violates (MC ↔ Ext$_x$) for all (possible) semantics returning conflict-free sets.

On the more general level, we see that asking for every conflict-free set to have a consistent base is very demanding. Roughly speaking, this is due the fact that attacks are binary whereas minimal conflicts may be ternary (or of a greater cardinality). Some authors argue that to obtain a consistent result, one should concentrate on admissibility and not on conflict-freeness. For example, Caminada and Vesic (2012) claim that $n$-ary attacks, for $n \geq 3$, are "simulated" in Dung's framework throughout the notion of admissibility. Thus, an idea for future work could be to study an alternative condition, which is that every *admissible* set has a consistent base.

### 4.2 Satisfying (MC↔Ext) and Different Acceptability Semantics

In this subsection, we study the properties related to particular semantics. We show that if an attack relation satisfies (MC ↔ Ext) for stable semantics, then it satisfies it for semi-stable semantics also. Then we identify conditions under which an attack relation satisfies (MC ↔ Ext) for stable semantics. We provide a similar result for preferred semantics. We also identify a sufficient condition so that an attack relation falsifies (MC ↔ Ext) under complete semantics. Then, we discuss the case of single-extension semantics, like grounded and ideal.

First, suppose that $\mathcal{R}$ satisfies (MC ↔ Ext$_s$). This means that for every finite set of formulae $\Sigma$, function Arg is a bijection between MC($\Sigma$) and Ext$_s$(Arg($\Sigma$), $\mathcal{R}$). Since every finite set of formulae has at least one maximal consistent subset (even if that is the empty set) then for every $\Sigma$, it must be that (Arg($\Sigma$), $\mathcal{R}$) has at least one stable extension. Since there are stable extensions, then stable and semi-stable semantics coincide (Caminada, 2006). Thus, we obtain the following proposition.

**Proposition 4.** *Let $\mathcal{R}$ be an attack relation. If $\mathcal{R}$ satisfies (MC ↔ Ext$_s$) then:*

- *for every finite set of formulae $\Sigma$ and $\mathcal{F} = (\text{Arg}(\Sigma), \mathcal{R})$, we have that $\text{Ext}_s(\mathcal{F}) = \text{Ext}_{ss}(\mathcal{F})$*

- *$\mathcal{R}$ satisfies (MC ↔ Ext$_{ss}$).*

Let us now prove that in the case of stable semantics, if the image with respect to Arg of every maximal consistent set is an extension and if the base of every extension is consistent, then the attack relation in question satisfies (MC ↔ Ext).

**Proposition 5.** *Let $\mathcal{R}$ be an attack relation. If for every set of formulae $\Sigma$ and $\mathcal{F} = (\text{Arg}(\Sigma), \mathcal{R})$, we have:*

- *for all $S \in \text{MC}(\Sigma)$, $\text{Arg}(S) \in \text{Ext}_s(\mathcal{F})$, and*

- *for all $\mathcal{E} \in \text{Ext}_s(\mathcal{F})$, $\text{Base}(\mathcal{E})$ is consistent*

*then $\mathcal{R}$ satisfies (MC ↔ Ext$_s$).*

*Proof.* Let us prove that $\mathcal{R}$ satisfies (MC ↔ Ext$_s$). We already know that for every $S \in \text{MC}(\Sigma)$, $\text{Arg}(S) \in \text{Ext}_s(\mathcal{F})$. Let us suppose that $\mathcal{E} \in \text{Ext}_s(\mathcal{F})$ and let us prove that there exists a unique set $S \in \text{MC}(\Sigma)$ such that $\text{Arg}(S) = \mathcal{E}$. If we prove that such a set exists, then uniqueness is guaranteed since for $S, S' \subseteq \Sigma$, if $S \neq S'$ then $\text{Arg}(S) \neq \text{Arg}(S')$ trivially holds. Thus, let us prove that there exists $S \in \text{MC}(\Sigma)$ such that $\text{Arg}(S) = \mathcal{E}$. Let $S' = \text{Base}(\mathcal{E})$ and let us prove that $S' \in \text{MC}(\Sigma)$ and $\text{Arg}(S') = \mathcal{E}$. By means of contradiction, suppose that $S'$ is a consistent but not maximal consistent set. Then, there exists $S'' \in \text{MC}(\Sigma)$ such that $S' \subseteq S''$. From the assumptions of this proposition, we have that $\mathcal{E}'' = \text{Arg}(S'')$ is a stable extension of $\mathcal{F}$. But we also have $\mathcal{E} \subsetneq \mathcal{E}''$. Since no stable extension is a proper subset of another stable extension, then $\mathcal{E}$ is not a stable extension. Contradiction. Thus, it must be that $S' \in \text{MC}(\Sigma)$. It is easy to see that $\mathcal{E} \subseteq \text{Arg}(\text{Base}(\mathcal{E}))$ (namely, for every set of arguments, by applying the function $\text{Arg}$ on its base, we obtain its superset). Let us prove $\text{Arg}(S') = \mathcal{E}$. Suppose the contrary. Then, $\mathcal{E} \subsetneq \text{Arg}(S')$. Since $S' \in \text{MC}(\Sigma)$ then $\text{Arg}(S') \in \text{Ext}_s(\mathcal{F})$. Thus, $\mathcal{E}$ is not a stable extension (since no stable extension is a proper subset of another stable extension). □

We prove that similar two conditions are sufficient to guarantee that $\mathcal{R}$ satisfies $(\text{MC} \leftrightarrow \text{Ext})$ under *preferred* semantics.

**Proposition 6.** *Let $\mathcal{R}$ be an attack relation. If for every set of formulae $\Sigma$ and $\mathcal{F} = (\text{Arg}(\Sigma), \mathcal{R})$, we have:*

- *for all $S \in \text{MC}(\Sigma)$, $\text{Arg}(S) \in \text{Ext}_p(\mathcal{F})$, and*
- *for all $\mathcal{E} \in \text{Ext}_p(\mathcal{F})$, $\text{Base}(\mathcal{E})$ is consistent*

*then $\mathcal{R}$ satisfies $(\text{MC} \leftrightarrow \text{Ext}_p)$.*

Proof of this property is similar to the proof of Proposition 5.

As a consequence of the two previous results, we can identify a sufficient condition so that $\mathcal{R}$ satisfies both $(\text{MC} \leftrightarrow \text{Ext}_s)$ and $(\text{MC} \leftrightarrow \text{Ext}_p)$.

**Corollary 5.** *Let $\mathcal{R}$ be an attack relation. If for every set of formulae $\Sigma$ and $\mathcal{F} = (\text{Arg}(\Sigma), \mathcal{R})$, we have:*

- *for all $S \in \text{MC}(\Sigma)$, $\text{Arg}(S) \in \text{Ext}_s(\mathcal{F})$, and*
- *for all $\mathcal{E} \in \text{Ext}_p(\mathcal{F})$, $\text{Base}(\mathcal{E})$ is consistent*

*then $\mathcal{R}$ satisfies both $(\text{MC} \leftrightarrow \text{Ext}_s)$ and $(\text{MC} \leftrightarrow \text{Ext}_p)$.*

*Proof.* Since every stable extension is a preferred one (Dung, 1995), it is clear that $\mathcal{R}$ satisfies both conditions of Proposition 5 and Proposition 6. By applying those propositions, we have that $\mathcal{R}$ satisfies $(\text{MC} \leftrightarrow \text{Ext}_s)$ and $(\text{MC} \leftrightarrow \text{Ext}_p)$. □

Let us now show that if an attack relation returns a stable extension having an inconsistent base, then it violates $(\text{MC} \leftrightarrow \text{Ext})$ for stable, semi-stable, preferred and complete semantics.

**Proposition 7.** *Let $\mathcal{R}$ be an attack relation. If there exists a finite set of formulae $\Sigma$ such that $\mathcal{F} = (\text{Arg}(\Sigma), \mathcal{R})$ has a stable extension $\mathcal{E}$ such that $\text{Base}(\mathcal{E})$ is inconsistent, then $\mathcal{R}$ falsifies $(\text{MC} \leftrightarrow \text{Ext}_x)$ for $x \in \{s, ss, p, c\}$.*

*Proof.* We supposed that there exists a stable extension $\mathcal{E} \in \text{Ext}_s(\mathcal{F})$ such that $\text{Base}(\mathcal{E}) \vdash \bot$. It has been proved (Dung, 1995) that every stable extension is a preferred and a complete one. We also know (Caminada, 2006) that $\mathcal{E}$ must be a semi-stable extension. By using Corollary 2, we conclude that $\mathcal{R}$ does not satisfy $(\text{MC} \leftrightarrow \text{Ext})$ for stable, semi-stable, preferred and complete semantics. □

Let us now study the case of complete semantics. We show that it is not possible for an attack relation to satisfy (MC ↔ $\text{Ext}_c$). The only condition we use in our result is that for every argument $a$, if $a$ has a formula $\varphi$ in its support, and $\neg\varphi \in \Sigma$, then there exists an argument $b \in \text{Arg}(\Sigma)$ such that $b$ attacks $a$.

**Proposition 8.** *Let $\mathcal{R}$ be an attack relation such that for every finite set of formulae $\Sigma$, for every $a \in \text{Arg}(\Sigma)$, for every $\varphi \in \text{Supp}(a)$, if there exists $\psi \in \Sigma$ such that $\psi \equiv \neg\varphi$ then there exists $b \in \text{Arg}(\Sigma)$ such that $(b, a) \in \mathcal{R}$. Then, $\mathcal{R}$ does not satisfy (MC ↔ $\text{Ext}_c$).*

*Proof.* We prove that if an attack relation $\mathcal{R}$ satisfies the condition from this proposition, then it falsifies (MC ↔ $\text{Ext}_c$). We use the proof by contradiction. In other words, the plan of the proof is as follows: first, we suppose that $\mathcal{R}$ satisfies the given condition. Second, by means of contradiction, we suppose that $\mathcal{R}$ *satisfies* (MC ↔ $\text{Ext}_c$). Third, we draw conclusions and obtain a contradiction. Fourth, by reductio ad absurdum, we conclude that it must be that $\mathcal{R}$ falsifies (MC ↔ $\text{Ext}_c$).

So, let us start by supposing the condition from the proposition *and* suppose that $\mathcal{R}$ satisfies (MC ↔ $\text{Ext}_c$). Thus, from Proposition 2, we obtain the $\mathcal{R}$ is conflict-dependent. Since $\mathcal{R}$ satisfies (MC ↔ $\text{Ext}_c$), then for every $\Sigma$, Arg is a bijection between MC($\Sigma$) and $\text{Ext}_c(\text{Arg}(\Sigma), \mathcal{R})$. Consider $\Sigma = \{\varphi, \neg\varphi, \psi\}$ and denote $\mathcal{F} = (\text{Arg}(\Sigma), \mathcal{R})$. It is clear that MC($\Sigma$) = $\{S_1, S_2\}$ with $S_1 = \{\varphi, \psi\}$ and $S_2 = \{\neg\varphi, \psi\}$. Since $\mathcal{R}$ satisfies (MC ↔ $\text{Ext}_c$) then $\text{Ext}_c(\mathcal{F}) = \{\mathcal{E}_1, \mathcal{E}_2\}$ with $\mathcal{E}_1 = \text{Arg}(S_1)$ and $\mathcal{E}_2 = \text{Arg}(S_2)$.

Let us now obtain a contradiction by proving that $\mathcal{E}_3 = \text{Arg}(\{\psi\})$ is a complete extension. First, prove that this set is conflict-free. Let $a, b \in \mathcal{E}_3$. Since $\mathcal{R}$ is conflict-dependent, then $(a, b) \notin \mathcal{R}$. Thus, $\mathcal{E}_3$ is conflict-free.

Let us now prove that for all $a \in \mathcal{E}_3$, for all $b \in \text{Arg}(\Sigma) \setminus \mathcal{E}_3$, we have that $(a, b) \notin \mathcal{R}$ and $(b, a) \notin \mathcal{R}$. By means of contradiction, suppose the contrary. Again from conflict-dependence, we have that $\text{Supp}(a) \cup \text{Supp}(b) \vdash \bot$. It must be that $\{\varphi, \neg\varphi\} \subseteq \text{Supp}(a) \cup \text{Supp}(b)$. Since the support of every argument is consistent, then $\text{Supp}(a)$ contains either $\varphi$ or $\neg\varphi$. Contradiction with the fact $a \in \text{Arg}(\{\psi\})$. Thus, $\mathcal{E}_3$ is an admissible set.

Let us now prove that $\mathcal{E}_3$ does not defend any arguments in $\text{Arg}(\Sigma) \setminus \mathcal{E}_3$. To show this, we only need to prove that every argument in $\text{Arg}(\Sigma) \setminus \mathcal{E}_3$ is attacked by at least one argument. Note that for every $a \in \text{Arg}(\Sigma) \setminus \mathcal{E}_3$, it holds that $a \in \mathcal{E}_1 \setminus \mathcal{E}_2$ or $a \in \mathcal{E}_2 \setminus \mathcal{E}_1$. Without loss of generality, let $a \in \mathcal{E}_1 \setminus \mathcal{E}_2$. Let us prove that $a$ is attacked. Note that in every argumentation system, every non-attacked argument is in all complete extensions. Since $a \notin \mathcal{E}_2$, then $a$ must be attacked. To sum up:

- $\mathcal{E}_3$ is an admissible set
- $\mathcal{E}_3$ does not attack any argument in $\text{Arg}(\Sigma) \setminus \mathcal{E}_3$
- $\text{Arg}(\Sigma) \setminus \mathcal{E}_3$ does not attack any argument in $\mathcal{E}_3$
- every argument in $\text{Arg}(\Sigma) \setminus \mathcal{E}_3$ is attacked by at least one argument.

Thus, $\mathcal{E}_3$ is a complete extension. Contradiction with the claim that $\text{Ext}_c(\mathcal{F}) = \{\mathcal{E}_1, \mathcal{E}_2\}$. By reductio ad absurdum, we conclude that $\mathcal{R}$ does not satisfy (MC ↔ $\text{Ext}_c$). □

What about the semantics which always return a unique extension, like grounded and ideal semantics? In such a case, it is not reasonable to expect that there is a bijection between MC($\Sigma$) and the set of extensions, since there can be several maximal consistent subsets of $\Sigma$. Let us formally state this fact.

**Proposition 9.** *If $x$ is a semantics such that for every argumentation system $\mathcal{F}$ we have $|\text{Ext}_x(\mathcal{F})| = 1$ then there is no attack relation $\mathcal{R}$ which satisfies $(\text{MC} \leftrightarrow \text{Ext}_x)$.*

*Proof.* Let $\Sigma = \{\varphi, \neg\varphi, \psi\}$. Denote $\mathcal{F} = (\text{Arg}(\Sigma), \mathcal{R})$. There are two maximal consistent subsets of $\Sigma$, i.e. $|\text{MC}(\Sigma)| = 2$. Since we supposed that every argumentation system has exactly one extension under semantics $x$, then there is no bijection between $\text{MC}(\Sigma)$ and $\text{Ext}_x(\mathcal{F})$. □

The previous simple result is not surprising. The idea between those semantics is to have one extension that contains all the arguments that should be accepted according to every point of view. Thus, we can expect a link between the set of formulae not belonging to any minimal inconsistent set and those extensions. Note that the sufficient conditions for $\mathcal{R}$ were identified (Gorogiannis & Hunter, 2011) so that for every finite set $\Sigma$ and $\mathcal{F} = (\text{Arg}(\Sigma), \mathcal{R})$ we have that the grounded and the ideal semantics coincide and that the extension is exactly $\text{Arg}(\Sigma \setminus (\Phi_1 \cup \ldots \cup \Phi_k))$ where $\{\Phi_1, \ldots, \Phi_k\}$ is the set of all minimal (for set inclusion) inconsistent subsets of $\Sigma$.

## 5. Identifying Classes of Attack Relations (Not-)Satisfying (MC↔Ext)

The previous sections show how to identify properties that an attack relation satisfying $(\text{MC} \leftrightarrow \text{Ext})$ must satisfy. They also provide several results closely related to the choice of a specific acceptability semantics. In this section, we identify classes of attack relations which satisfy, do not satisfy $(\text{MC} \leftrightarrow \text{Ext})$, or serve as lower (upper) bounds (with respect to set inclusion) for (non-)satisfying $(\text{MC} \leftrightarrow \text{Ext})$.

We first show that the whole class of *symmetric* attack relations *violates* $(\text{MC} \leftrightarrow \text{Ext})$ for all semantics from Definition 4.

**Proposition 10.** *If $\mathcal{R}$ is a symmetric attack relation, then for every $x \in \{s, ss, p, c, g, i\}$, $\mathcal{R}$ falsifies $(\text{MC} \leftrightarrow \text{Ext}_x)$.*

*Proof.* From Proposition 9, we see that $\mathcal{R}$ violates $(\text{MC} \leftrightarrow \text{Ext}_g)$ and $(\text{MC} \leftrightarrow \text{Ext}_i)$. Let us now prove the same for other acceptability semantics.

If $\mathcal{R}$ is a symmetric attack relation, then every conflict-free set is admissible. Furthermore, it is easy to see that in this case every maximal conflict-free set is a stable extension (and vice versa). Since every finite argumentation system has at least one maximal conflict-free set, then every finite argumentation system using a symmetric attack relation has at least one stable extension. Let $\mathcal{R}$ be a symmetric relation and suppose that for at least one $x \in \{s, ss, p, c\}$, $\mathcal{R}$ satisfies $(\text{MC} \leftrightarrow \text{Ext}_x)$. From Corollary 4, we conclude that $\mathcal{R}$ is not valid. This means that there exists a finite propositional knowledge base $\Sigma$ and $\mathcal{F} = (\text{Arg}(\Sigma), \mathcal{R})$ such that there is a conflict-free set $\mathcal{E} \subseteq \text{Arg}(\Sigma)$ having an inconsistent base. Let $\mathcal{E}' \subseteq \text{Arg}(\Sigma)$ be a maximal conflict-free set containing $\mathcal{E}$, i.e. such that $\mathcal{E} \subseteq \mathcal{E}'$. Since $\mathcal{E}'$ is a maximal conflict-free set, then it is a stable extension of $\mathcal{F}$. Since $\mathcal{E}'$ is a stable extension, then it is also a semi-stable, preferred and a complete one. Since $\text{Base}(\mathcal{E}') \vdash \bot$ then Corollary 2 implies that for all $x \in \{s, ss, p, c\}$, $\mathcal{R}$ fails to satisfy $(\text{MC} \leftrightarrow \text{Ext}_x)$. □

We now identify another class of attack relations that do not satisfy $(\text{MC} \leftrightarrow \text{Ext})$. Namely, we show that every (possible) attack generating "too many attacks" falsifies $(\text{MC} \leftrightarrow \text{Ext})$. First, we need to formally define what we mean by "too many attacks". We do this by introducing the notion of conflict-completeness.

**Definition 8** (Conflict-complete). *Let $\mathcal{R} \subseteq \text{Arg}(\mathcal{L}) \times \text{Arg}(\mathcal{L})$ be an attack relation. We say that $\mathcal{R}$ is conflict-complete if and only if for every minimal conflict $C \subseteq \mathcal{L}$ (i.e. for every inconsistent set whose every proper subset is consistent), for every $C_1, C_2 \subseteq C$ such that $C_1 \neq \emptyset$, $C_2 \neq \emptyset$, $C_1 \cup C_2 = C$, for every argument $a_1$ such that $\text{Supp}(a_1) = C_1$, there exists an argument $a_2$ such that $\text{Supp}(a_2) = C_2$ and $(a_2, a_1) \in \mathcal{R}$.*

Intuitively, an attack relation is *conflict-complete* if when two sets form a minimal conflict, then every argument built from one of the two sets can be attacked by an argument from the other set. This notion is inspired by the desire to describe properties of a class of existing (and new) attack relations. For example, *canonical undercut* is conflict-complete.

We can show that if an attack relation is *conflict-complete*, then it *falsifies* (MC ↔ Ext) for stable, semi-stable, preferred and complete semantics.

**Proposition 11.** *Let $\mathcal{R}$ be an attack relation. If $\mathcal{R}$ is conflict-complete then $\mathcal{R}$ does not satisfy (MC ↔ $\text{Ext}_x$) for $x \in \{s, ss, p, c\}$.*

*Proof.* Let $\mathcal{R}$ be a conflict-complete attack relation and let us use the proof by contradiction. Thus, suppose that there exist $x \in \{s, ss, p, c\}$ such that $\mathcal{R}$ satisfies (MC ↔ $\text{Ext}_x$) and obtain a contradiction. From Proposition 2, we have that $\mathcal{R}$ is conflict-dependent. Let $\Sigma = \{\varphi, \varphi \to \psi, \neg\psi\}$. Let $\mathcal{F} = (\text{Arg}(\Sigma), \mathcal{R})$ and $\mathcal{E} = \text{Arg}(\{\varphi\}) \cup \text{Arg}(\{\varphi \to \psi\}) \cup \text{Arg}(\{\neg\psi\})$. We prove that $\mathcal{E}$ is a stable extension of $\mathcal{F}$. First, prove that $\mathcal{E}$ is conflict-free. Let $a, b \in \mathcal{E}$ and suppose $(a, b) \in \mathcal{R}$. From conflict-dependence, we obtain $\text{Supp}(a) \cup \text{Supp}(b) \vdash \bot$. Contradiction with the definition of $\mathcal{E}$, since there are no two arguments of $\mathcal{E}$ such that the union of their supports is inconsistent. Now, prove that $\mathcal{E}$ attacks every argument in $\text{Arg}(\Sigma) \setminus \mathcal{E}$. Let $a' \in \text{Arg}(\Sigma) \setminus \mathcal{E}$. There are three cases. Case 1: $\text{Supp}(a') = \{\varphi, \varphi \to \psi\}$. In this case, since $\mathcal{R}$ is conflict-complete, then $a'$ is attacked by at least one argument from the set $\text{Arg}(\neg\psi)$. Case 2: $\text{Supp}(a') = \{\varphi, \neg\psi\}$. Again from conflict-completeness, such an argument is attacked by an argument from the set $\text{Arg}(\{\varphi \to \psi\})$. Case 3: $\text{Supp}(a') = \{\varphi \to \psi, \neg\psi\}$ is also similar, since $a'$ is then attacked by an argument having support $\{\varphi\}$. We conclude that $\mathcal{E} \in \text{Ext}_s(\mathcal{F})$. It is easy to see that $\text{Base}(\mathcal{E}) \vdash \bot$. Proposition 7 now implies that $\mathcal{R}$ does not satisfy (MC ↔ $\text{Ext}_x$) for every $x \in \{s, ss, p, c\}$. Contradiction. □

The previous part of this paper studies classes of attack relations. Let us now define some particular cases of attack relations. If $\Phi = \{\varphi_1, \ldots, \varphi_k\}$ is a set of formulae, notation $\bigwedge \Phi$ stands for $\varphi_1 \wedge \ldots \wedge \varphi_k$.

**Definition 9** (Attack relations). *Let $a, b \in \text{Arg}(\mathcal{L})$. We define the following attack relations:*

- *defeat: $a\mathcal{R}_d b$ if and only if $\text{Conc}(a) \vdash \bigwedge \neg\text{Supp}(b)$*

- *direct defeat: $a\mathcal{R}_{dd} b$ if and only if there exists $\varphi \in \text{Supp}(b)$ such that $\text{Conc}(a) \vdash \neg\varphi$*

- *undercut: $a\mathcal{R}_u b$ if and only if there exists $\Phi \subseteq \text{Supp}(b)$ such that $\text{Conc}(a) \equiv \neg \bigwedge \Phi$*

- *direct undercut: $a\mathcal{R}_{du} b$ if and only if there exists $\varphi \in \text{Supp}(b)$ such that $\text{Conc}(a) \equiv \neg\varphi$*

- *canonical undercut: $a\mathcal{R}_{cu} b$ if and only if $\text{Conc}(a) \equiv \neg \bigwedge \text{Supp}(b)$*

- rebut: $a\mathcal{R}_r b$ *if and only if* $\text{Conc}(a) \equiv \neg\text{Conc}(b)$

- defeating rebut: $a\mathcal{R}_{dr} b$ *if and only if* $\text{Conc}(a) \vdash \neg\text{Conc}(b)$

- conflicting attack: $a\mathcal{R}_c b$ *if and only if* $\text{Supp}(a) \cup \text{Supp}(b) \vdash \bot$

- rebut + direct undercut: $a\mathcal{R}_{rdu} b$ *if and only if* $a\mathcal{R}_r b$ or $a\mathcal{R}_{du} b$

- big argument attack: $a\mathcal{R}_{ba} b$ *if and only if there exists* $\varphi \in \text{Supp}(b)$ *such that* $\text{Supp}(a) \vdash \neg\varphi$.

The first seven items from the previous definition list are, to the best of our knowledge, all the attack relations used in the logic-based argumentation literature. Finding the exact paper in which each of them occurs for the first time would be quite a challenging task. We can say that *rebut* was defined by Pollock (1987, 1992). *Direct undercut* was introduced in the work of Elvang-Gøransson, Fox, and Krause (1993) and Elvang-Gøransson and Hunter (1995). *Undercut* and *canonical undercut* were defined in this form by Besnard and Hunter (2000, 2001). To the best of our knowledge, *conflicting attack* was not used in the argumentation literature. A possibility to use such a relation was mentioned (Besnard & Hunter, 2008, p. 35). We show that it is "not enough" to capture the presence of inconsistency to make a good attack relation. Namely, we show later that this attack relation may return inconsistent extensions. *Rebut + direct undercut* is added by the author of the present paper, as an attempt to investigate the possibility to use *rebut* to detect some conflicts not detected by *direct undercut*, but to avoid using a symmetric relation (rebut). The name *big argument attack* and the idea behind this attack relation are due to L. van der Torre (personal communication, June 18, 2012). This attack relation was coined with the goal to show that there are reasonable attack relations not taking into account the conclusion of an argument. We later show (Proposition 16) that this attack relation also satisfies ($\text{MC} \leftrightarrow \text{Ext}$). (The idea behind the name of this attack relation is that it is sufficient to use only one argument per support since the conclusions are not important. Those arguments are called *big* since one big argument plays a role of a whole class of "normal" arguments, i.e. all the arguments having the same support. The attack relation is called *big* since it is to be used between big arguments.)

The reader can easily check that *canonical undercut* is conflict-complete, which leads to the conclusion that every attack relation containing canonical undercut (in the set-theoretic sense) is also conflict-complete.

**Proposition 12.** *Let* $\mathcal{R} \subseteq \text{Arg}(\mathcal{L}) \times \text{Arg}(\mathcal{L})$ *be an attack relation. If* $\mathcal{R}_{cu} \subseteq \mathcal{R}$ *then* $\mathcal{R}$ *is conflict-complete.*

Thus, from Proposition 11, we conclude that every attack relation containing *canonical undercut* falsifies ($\text{MC} \leftrightarrow \text{Ext}$) for stable, semi-stable, preferred and complete semantics.

**Corollary 6.** *Let* $\mathcal{R}$ *be an attack relation. If* $\mathcal{R}_{cu} \subseteq \mathcal{R}$, *then* $\mathcal{R}$ *does not satisfy* ($\text{MC} \leftrightarrow \text{Ext}_x$) *for* $x \in \{s, ss, p, c\}$.

Since $\mathcal{R}_{cu} \subseteq \mathcal{R}_u \subseteq \mathcal{R}_d \subseteq \mathcal{R}_c$, then we obtain that as soon as an attack relation $\mathcal{R}$ contains $\mathcal{R}_u$, or $\mathcal{R}_d$ or $\mathcal{R}_c$ then it falsifies ($\text{MC} \leftrightarrow \text{Ext}$) for stable, semi-stable, preferred and complete semantics.

**Corollary 7.** *Let* $\mathcal{R}$ *be an attack relation. If* $\mathcal{R}_u \subseteq \mathcal{R}$, *or* $\mathcal{R}_d \subseteq \mathcal{R}$ *or* $\mathcal{R}_c \subseteq \mathcal{R}$ *then* $\mathcal{R}$ *falsifies* ($\text{MC} \leftrightarrow \text{Ext}_x$) *for* $x \in \{s, ss, p, c\}$.

Hence, there is a whole class of attack relations based on undercutting which do not satisfy (MC ↔ Ext). We also identified another class of attack relations, this time based on rebutting, which do not satisfy (MC ↔ Ext$_s$). Namely, every attack relation contained in defeating rebut must falsify (MC ↔ Ext$_s$). Observe how the proof of the following proposition is based on the idea from Example 3.

**Proposition 13.** *Let $\mathcal{R}$ be an attack relation. If $\mathcal{R} \subseteq \mathcal{R}_{dr}$ then $\mathcal{R}$ does not satisfy (MC ↔ Ext$_s$).*

*Proof.* Let us suppose the contrary, i.e. let $\mathcal{R}$ satisfy (MC ↔ Ext$_s$). Let $\Sigma = \{\varphi \wedge \psi, \varphi \wedge \neg\psi\}$ and $\mathcal{F} = (\text{Arg}(\Sigma), \mathcal{R})$. We have that $\text{MC}(\Sigma) = \{S_1, S_2\}$, with $S_1 = \{\varphi \wedge \psi\}$ and $S_2 = \{\varphi \wedge \neg\psi\}$. Thus, it must be that $\text{Ext}_s(\mathcal{F}) = \{\mathcal{E}_1, \mathcal{E}_2\}$ with $\mathcal{E}_1 = \text{Arg}(S_1)$ and $\mathcal{E}_2 = \text{Arg}(S_2)$. It is obvious that for an argument $a_1 = (\{\varphi \wedge \psi\}, \varphi \vee \psi)$ we must have $a_1 \in \mathcal{E}_1$. Since $\mathcal{E}_2$ is a stable extension, then there must exist an argument $a_2 \in \mathcal{E}_2$ such that $(a_2, a_1) \in \mathcal{R}$. Thus, it must be that $\text{Conc}(a_2) \vdash \neg\varphi \wedge \neg\psi$. Consequently, $\text{Conc}(a_2) \vdash \neg\varphi$. Recall that $\text{Supp}(a_2) = \{\varphi \wedge \psi\}$ or $\text{Supp}(a_2) = \{\varphi \wedge \neg\psi\}$. Contradiction. □

Since $\mathcal{R}_r \subseteq \mathcal{R}_{dr}$ then the previous conclusion holds for every relation contained in $\mathcal{R}_r$.

**Corollary 8.** *Let $\mathcal{R}$ be an attack relation. If $\mathcal{R} \subseteq \mathcal{R}_r$ then $\mathcal{R}$ does not satisfy (MC ↔ Ext$_s$).*

*Proof.* Let $\mathcal{R} \subseteq \mathcal{R}_r$. Since $\mathcal{R}_r \subseteq \mathcal{R}_{dr}$, then $\mathcal{R} \subseteq \mathcal{R}_{dr}$. From Proposition 13, $\mathcal{R}$ falsifies (MC ↔ Ext$_s$). □

## 6. Particular Attack Relations and (MC↔Ext)

In the previous section, we identified classes of relations which do not satisfy (MC ↔ Ext). In this section, we examine in detail all the attack relations from Definition 9.

By using the results presented until now, we prove that *direct undercut*, *direct defeat* and *big argument attack* satisfy (MC ↔ Ext) for stable, semi-stable and preferred semantics, and falsify it for other semantics, whereas other attack relations fail to satisfy (MC ↔ Ext) for any semantics.

Note that it has been proved (Cayrol, 1995) that direct undercut satisfies (MC ↔ Ext) in the case of stable semantics. From Proposition 4, we conclude that direct undercut satisfies (MC ↔ Ext) for semi-stable semantics. So, we only need to prove that $\mathcal{R}_{du}$ satisfies (MC ↔ Ext) in the case of preferred semantics.

**Proposition 14.** *Attack relation $\mathcal{R}_{du}$ satisfies (MC ↔ Ext$_x$) for $x \in \{s, ss, p\}$.*

*Proof.* We have already seen why $\mathcal{R}_{du}$ satisfies (MC ↔ Ext) under stable and semi-stable semantics. We now study the case of preferred semantics. Let $\Sigma$ be a finite set of formulae and $\mathcal{F} = (\text{Arg}(\Sigma), \mathcal{R}_{du})$. Since it was already proved (Cayrol, 1995) that stable extensions of $\mathcal{F}$ are exactly $\text{Arg}(S)$, when $S$ ranges over $\text{MC}(\Sigma)$, and since every stable extension is a preferred one, then it is clear that for every $S \in \text{MC}(\Sigma)$ we have that $\text{Arg}(S)$ is a preferred extension of $\mathcal{F}$. Thanks to Proposition 6, we now only need to prove that the base of every preferred extension is consistent. This result follows from Prop. 34 by Gorogiannis and Hunter (2011), since relation $\mathcal{R}_{du}$ satisfies all the conditions of that proposition. Thus, direct undercut satisfies (MC ↔ Ext$_p$). □

**Example 6.** *Consider a relation $\rightsquigarrow$ for inferring from an inconsistent knowledge base defined as follows: given a set $\Sigma$, we write $\Sigma \rightsquigarrow \varphi$ if and only if for every maximal consistent subset $S$ of $\Sigma$, it* holds that $S \vdash \varphi$, where $\vdash$ stands for classical entailment. Now, consider an argumentation system using *direct undercut* as attack relation and stable semantics. From Proposition 14, we conclude that for every $\Sigma$, Arg *is a bijection between* $\mathtt{MC}(\Sigma)$ *and* $\mathtt{Ext}_s(\mathtt{Arg}(\Sigma), \mathcal{R}_{du})$. *Roughly speaking, this means that the set of formulae that can be inferred from $\Sigma$ with respect to $\rightsquigarrow$ is equal to the set of formulae that are conclusions of all the extensions of the corresponding argumentation framework based on direct undercut and stable semantics. More formally: for every $\Sigma \subseteq \mathcal{L}$, for every formula $\varphi \in \mathcal{L}$ we have that:* $\Sigma \rightsquigarrow \varphi$ *if and only if for every extension* $\mathcal{E} \in \mathtt{Ext}_s(\mathtt{Arg}(\Sigma), \mathcal{R}_{du})$, *there exists* $a \in \mathcal{E}$ *such that* $\varphi = \mathtt{Conc}(a)$.

Let us now show that $\mathcal{R}_{dd}$ also satisfies (MC ↔ Ext) for stable, semi-stable and preferred semantics.

**Proposition 15.** *Attack relation* $\mathcal{R}_{dd}$ *satisfies* (MC ↔ $\mathtt{Ext}_x$) *for* $x \in \{s, ss, p\}$.

*Proof.* Let $\Sigma$ be a finite knowledge base and $\mathcal{F} = (\mathtt{Arg}(\Sigma), \mathcal{R}_{dd})$. Let $S \in \mathtt{MC}(\Sigma)$. It is easy to see that $\mathcal{E} = \mathtt{Arg}(S)$ is a stable extension of $\mathcal{F}$. Namely, $\mathcal{E}$ is conflict-free since $\mathcal{R}_{dd}$ is conflict-dependent. Furthermore, for every argument $a' \in \mathtt{Arg}(\Sigma) \setminus \mathcal{E}$, it must be that $\mathtt{Supp}(a')$ contains at least one formulae $\varphi \in \Sigma \setminus S$. From this fact, it is easy to conclude that there exists an argument $a \in \mathcal{E}$ such that $\mathtt{Supp}(a) \subseteq S$ and $\mathtt{Conc}(a) \equiv \neg\varphi$ (since $S$ is a maximal consistent set). Thus, $a$ attacks $a'$ which ends this part of the proof and shows why $\mathcal{E} \in \mathtt{Ext}_s(\mathcal{F})$. Since every stable extension is a semi-stable and a preferred one, then $\mathcal{E} \in \mathtt{Ext}_{ss}(\mathcal{F})$ and $\mathcal{E} \in \mathtt{Ext}_p(\mathcal{F})$. Let us now suppose that $\mathcal{E}$ is a preferred extension of $\mathcal{F}$. Since direct defeat satisfies conditions of Prop. 34 by Gorogiannis and Hunter (2011), then we conclude that $\mathtt{Base}(\mathcal{E})$ is consistent. From Corollary 5, we conclude that $\mathcal{R}_{dd}$ satisfies (MC ↔ Ext) for stable and preferred semantics. Now, Proposition 4 implies that $\mathcal{R}_{dd}$ also satisfies (MC ↔ $\mathtt{Ext}_{ss}$). □

We now show that it is not necessary to look at conclusions of arguments in order to satisfy (MC ↔ Ext). Namely, we can show that *big argument attack* satisfies (MC ↔ Ext) for stable, semi-stable and preferred semantics.

**Proposition 16.** *Attack relation* $\mathcal{R}_{ba}$ *satisfies* (MC ↔ $\mathtt{Ext}_x$) *for* $x \in \{s, ss, p\}$.

*Proof.* Let us first show that for every $S \in \mathtt{MC}(\Sigma)$, it holds that $\mathcal{E} = \mathtt{Arg}(S)$ is a stable extension in $\mathcal{F} = (\mathtt{Arg}(\Sigma), \mathcal{R}_{ba})$. Since $\mathcal{R}_{ba}$ is conflict-dependent, then $\mathcal{E}$ is conflict-free. Let $a' \in \mathtt{Arg}(\Sigma) \setminus \mathcal{E}$ and let us prove that there exists $a \in \mathcal{E}$ such that $a\mathcal{R}_{ba}a'$. Since $a' \notin \mathcal{E}$, then there exists $\varphi \in \mathtt{Supp}(a')$ such that $\varphi \notin S$. Since $S$ is a maximal consistent subset of $\Sigma$, then $S \vdash \neg\varphi$. Let $S' \subseteq S$ be a minimal with respect to set inclusion consistent set such that $S' \vdash \neg\varphi$ (such a set exists since $S$ is consistent) and let $a = (S', \neg\varphi)$. $a$ is an argument since $S'$ is a minimal consistent set from which $\neg\varphi$ can be deduced. We see that $(a, a') \in \mathcal{R}_{ba}$. This means that the image with respect to Arg of every maximal consistent subset of $\Sigma$ is a stable extension of $\mathcal{F}$. Thus, it is also a semi-stable and a preferred extension of $\mathcal{F}$. Let us now prove that for every $\Sigma$ and the corresponding $\mathcal{F} = (\mathtt{Arg}(\Sigma), \mathcal{R}_{ba})$, the base of every preferred extension $\mathcal{E}$ of $\mathcal{F}$ is a consistent set. Let $\mathcal{E} \in \mathtt{Ext}_p(\mathcal{F})$ and $S = \mathtt{Base}(\mathcal{E})$. Aiming to a contradiction, suppose the contrary, i.e. let $S$ be an inconsistent set. Let $S' \subseteq S$ be a minimal (with respect to set inclusion) inconsistent set. Denote $S' = \{\varphi_1, \ldots, \varphi_n\}$. Let $a \in \mathcal{E}$ be an argument such that $\varphi_n \in \mathtt{Supp}(a)$, and let $a' = (S' \setminus \{\varphi_n\}, \neg\varphi_n)$. It is clear that $(a', a) \in \mathcal{R}_{ba}$. Since $\mathcal{E}$ is a preferred extension, it is conflict-free, thus $a' \notin \mathcal{E}$. Furthermore, $\mathcal{E}$ is admissible, so there must exist $b \in \mathcal{E}$ such that $(b, a) \in \mathcal{R}_{ba}$. Since $(b, a) \in \mathcal{R}_{ba}$, then there exists $i \in \{1, \ldots, n-1\}$ such that $\text{Supp}(b) \vdash \neg\varphi_i$. Since $\varphi_i \in S'$, then there exists an argument $c \in \mathcal{E}$ such that $\varphi_i \in \text{Supp}(c)$. According to the definition of $\mathcal{R}_{ba}$, that would mean that $b$ attacks $c$. Contradiction with the fact that $\mathcal{E}$ is conflict-free. So, $S$ must be a consistent set. This shows that for every $\Sigma$ and the corresponding $\mathcal{F} = (\text{Arg}(\Sigma), \mathcal{R}_{ba})$, the base of every preferred extension $\mathcal{E}$ of $\mathcal{F}$ is a consistent set. From Corollary 5, we conclude that $\mathcal{R}_{ba}$ satisfies ($\text{MC} \leftrightarrow \text{Ext}$) for stable and preferred semantics. Now, Proposition 4 implies that $\mathcal{R}_{ba}$ also satisfies ($\text{MC} \leftrightarrow \text{Ext}_{ss}$). □

We already know that no relation satisfies ($\text{MC} \leftrightarrow \text{Ext}$) for the grounded or ideal semantics. By using Proposition 8, it is easy to conclude that $\mathcal{R}_{du}$, $\mathcal{R}_{dd}$ and $\mathcal{R}_{ba}$ falsify ($\text{MC} \leftrightarrow \text{Ext}_c$).

Let us now prove that the remaining attack relations from Definition 9 do not satisfy ($\text{MC} \leftrightarrow \text{Ext}$) for neither of semantics from Definition 4.

**Proposition 17.** *Attack relations $\mathcal{R}_d$, $\mathcal{R}_u$, $\mathcal{R}_{cu}$, $\mathcal{R}_r$, $\mathcal{R}_{dr}$, $\mathcal{R}_{rdu}$, $\mathcal{R}_c$ falsify* ($\text{MC} \leftrightarrow \text{Ext}$) *for stable, semi-stable, preferred, complete, grounded and ideal semantics.*

*Proof.* Note that we already showed that no attack relation satisfies ($\text{MC} \leftrightarrow \text{Ext}$) for grounded or ideal semantics. So, in the rest of the proof, we only need to consider stable, semi-stable, preferred and complete semantics.

Let us first consider the attack relations $\mathcal{R}_{cu}$, $\mathcal{R}_u$, $\mathcal{R}_d$ and $\mathcal{R}_c$. By using Proposition 11, we conclude that those relations violate ($\text{MC} \leftrightarrow \text{Ext}$) for stable, semi-stable, preferred and complete semantics.

It is obvious that relations $\mathcal{R}_r$ and $\mathcal{R}_c$ are symmetric. Note that $\mathcal{R}_{dr}$ is also symmetric: this comes from the fact that $\varphi \vdash \neg\psi$ if and only if $\varphi, \psi \vdash \bot$ if and only if $\psi \vdash \neg\varphi$. Thus, Proposition 10 yields a conclusion that they do not satisfy ($\text{MC} \leftrightarrow \text{Ext}$) for neither of the considered acceptability semantics.

Let us now study the relation $\mathcal{R}_{rdu}$. Let $\Sigma = \{\varphi, \varphi \to \psi, \neg\psi\}$ and $\mathcal{F} = (\text{Arg}(\Sigma), \mathcal{R}_{rdu})$. Let us define a set $\mathcal{E}$ of arguments as follows: $\mathcal{E} = \{a \in \text{Arg}(\Sigma) \mid \text{Conc}(a) \not\equiv \neg\varphi \text{ and } \text{Conc}(a) \not\equiv \neg(\varphi \to \psi) \text{ and } \text{Conc}(a) \not\equiv \psi\}$.

Prove that $\mathcal{E}$ is conflict-free. Let $a, b \in \mathcal{E}$ and let $a\mathcal{R}_{rdu}b$. Whether $a\mathcal{R}_r b$ or $a\mathcal{R}_{du}b$ is not important, since in both cases, we obtain $\text{Supp}(a) \cup \text{Supp}(b) \vdash \bot$. Contradiction, since there are no two formulae in $\Sigma$ whose union is an inconsistent set. So $\mathcal{E}$ is a conflict-free set.

Suppose that $a' \in \text{Arg}(\Sigma) \setminus \mathcal{E}$. So, $\text{Conc}(a') \equiv \neg\varphi$ or $\text{Conc}(a') \equiv \neg(\varphi \to \psi)$ or $\text{Conc}(a') \equiv \psi$. In any of those cases, $a'$ is attacked by at least one argument from $\mathcal{E}$, namely by $(\{\varphi\}, \varphi)$, or by $(\{\varphi \to \psi\}, \varphi \to \psi)$, or by $(\{\neg\psi\}, \neg\psi)$. So, $\mathcal{E}$ is a stable extension, and consequently, semi-stable, preferred and complete extension. It is obvious that $\text{Base}(\mathcal{E})$ is an inconsistent set, so by Corollary 2 we conclude that $\mathcal{R}_{rdu}$ does not satisfy ($\text{MC} \leftrightarrow \text{Ext}$) for stable, semi-stable, preferred and complete semantics. □

## 7. Discussion, Related and Future Work

This paper identified and studied the large class of instantiations of Dung's abstract theory corresponding to the *maxi-consistent operator*. In other words, we studied the instantiations where every extension of the argumentation system corresponds to exactly one maximal consistent subset of the knowledge base. We proved properties of attack relations belonging to this class: they must be conflict-dependent, must not be valid, must not be conflict-complete, must not be symmetric etc. We also identified some attack relations serving as lower or upper bounds of the class. By using our results, we showed for all existing attack relations from the argumentation literature whether or not they belong to this class. We also showed for the first time that an attack relation not depending on arguments' conclusions can return reasonable results. Furthermore, we showed that such a relation is a member of (MC ↔ Ext) class.

Practical benefits of the work reported in this paper, and more generally, any work devoted to studying the link between a class of instantiations of Dung's theory and an operator, can be resumed as follows.

**(I)** A case when an instantiation of Dung's theory is shown to correspond to an existing operator.

First, such a work can help to "validate" an argumentation-based approach by showing in which cases it returns a result comparable with that of a non argumentation-based approach. The possible criticism of such an instantiation is that it is useless, since one can obtain the same result without using argumentation. But, this is far from being true; namely, argumentation can be used for explanatory purposes. For example, if one wants to know why a certain conclusion is accepted, an argument having that conclusion can be presented. That argument can be attacked by other arguments and so on. Also, it might be possible to construct only a *part of the argumentation graph* related to the argument in question, thus having a better knowledge representation (i.e. ignoring the parts of the knowledge base unrelated to the argument one wants to concentrate on).

The second benefit of this type of work is that it can help to *reduce computational complexity* by using the simpler approach in the cases when the result obtained by an argumentation-based approaches and a non argumentation-based approaches is the same. Please note that the work in this category (capturing an operator with an instantiation of Dung's theory) is far from being limited to the case of the maxi-consistent operator, as it was shown by Vesic and van der Torre (2012) that there exists a large class of instantiations of the abstract argumentation theory returning a consistent result substantially different from the one returned by the maxi-consistent operator.

**(II)** A case when an instantiation of Dung's theory does not correspond to any existing operator.

Working on the links between instantiations of Dung's theory and operators can be even more beneficial in the case when an instantiation of the abstract argumentation theory does not corresponding to any known operator happens to be found. We distinguish three possible situations.

(a) A case when an instantiation calculates a "useful" result which can be obtained by an operator, but that operator was unknown until now. In such a case, a new operator is discovered thanks to argumentation. The question is then, in which situations to use argumentative approach, and when to apply the operator? The answer depends on the balance between the need for computational efficiency (which we conjecture is often on the side of the approach directly applying the operator) and the need to represent knowledge in a format that is easy to grasp, argue and justify an accepted piece of knowledge, which are the usual advantages of argumentation.

(b) A case when an instantiation of Dung's abstract theory returns a "useful" result which cannot be obtained by any operator. Recall that an operator is a function that, for every finite knowledge base, returns a set of its subsets. But, an argumentative approach could return a result that cannot be represented in that form, for instance, if an argument $(\Phi, \alpha)$ is in an extension, whereas $(\Phi, \beta)$ is not, with $\alpha \neq \beta$. Thus, the expressive power of the operator-based approach might be not enough to distinguish those subtleties. A very important question of how to define such an instantiation is still open. Another relevant issue is to see in which context such instantiations make sense and how they can be applied.

(c) A case when an instantiation returns a "bad" result. This class regroups a set of instantiations representing a behaviour one would like to avoid. The general question: "how to distinguish *useful* from *bad* instantiations?" is certainly a hard one. Apart from a scientific debate, evaluation can include tests on a set of benchmark examples. Note that the limits of testing a reasoning formalism on a set of benchmark examples have been pointed out by Vreeswijk (1995). Another, more principled (and more demanding) way to proceed is to define a set of postulates to be satisfied by an argumentation formalism (Caminada & Amgoud, 2007; Caminada, Carnielli, & Dunne, 2012).

As a remark, note that the fact that an instantiation may return an inconsistent result, does not mean that it is completely useless. Namely, there might be cases when arguments are constructed from an inconsistent knowledge base, and when one resolves just some of the existing inconsistencies by an argumentative approach, and *then* applies another inconsistency-tolerant approach. Also, inconsistency handling is not the only use of argumentation. Thus, still in the same setting, a drastic case would be to first use argumentation for another purpose (not dealing with at all with inconsistencies) and then apply a different approach to reason with inconsistency.

We now review the related work. Maxi-consistent sets play a major role in the characterization of various forms of non-classical logical reasoning (Bochman, 2001) and in belief revision (Alchourrón, Gardenfors, & Makinson, 1985). The remainder of the section considers the papers having a link with argumentation.

The paper by Cayrol (1995) is one of the early works relating the results obtained directly from a knowledge base and by using an argumentative approach. In that paper, it was shown that *direct undercut* satisfies (MC ↔ Ext) for stable semantics, but no results for other semantics or attack relations were provided. We not only studied other attack relations and other semantics, but also provided a general study of properties an attack relation satisfying (MC ↔ Ext) must also satisfy.

Amgoud and Vesic (2010) generalised the result by Cayrol (1995) for the case of prioritised knowledge base, by showing that Arg is a bijection between preferred sub-theories (Brewka, 1989), which generalise maximal consistent sets in case of prioritised knowledge base, and stable extensions of the corresponding preference-based argumentation system using *direct undercut* as an attack relation and the *weakest link principle* as a preference relation.

Amgoud and Besnard (2009, 2010) also studied the link between a knowledge base and the corresponding argumentation system. Those papers introduced some important notions like conflict-dependence and validity of an attack relation and proved numerous results related to consistency in the underlying logic. However, note that the criterion (MC ↔ Ext) was neither defined nor studied in those papers; they provided (Amgoud & Besnard, 2010, Corollary 1) a link between MC($\Sigma$) and maximal conflict-free sets of $\mathcal{F} = (\text{Arg}(\Sigma), \mathcal{R})$. Furthermore, this result is proved under hypotheses which are impossible to satisfy: the attack relation should be both valid and conflict-dependent, which is impossible (as proved in Proposition 3). Some other results in that paper (Amgoud & Besnard, 2010, e.g. Prop. 4) are proved only for an attack relation which is both conflict-dependent and conflict-sensitive, which is not the case for any of the well-known attack relations. Consequently, the majority of the negative results of those papers are only applicable to a minority of attack relations. Furthermore, examples in those papers are often "incomplete systems"; thus, it is not surprising that there is no link between MC(Base($\mathcal{A}$)) and Ext($\mathcal{A}, \mathcal{R}$) in those examples.

A recent paper by Gorogiannis and Hunter (2011) studied the properties of attack relations in the case when a Dung-style argumentation system is instantiated with classical propositional logic. Our work is related to those ideas, however, the focus of our paper is different. Our main goal is to study to which extent Dung's theory can be used as a general framework for reasoning.

On the technical side, we concentrate on studying the properties of the *class* of attack relations satisfying (MC ↔ Ext) and identifying attack relations serving as lower and upper bounds of classes of relations non-satisfying (MC ↔ Ext).

One of the open questions is to find a set of conditions such that an attack relation satisfies those conditions if and only if it satisfies (MC ↔ Ext). Until recently, *direct undercut* and *direct defeat* were the only known attack relations satisfying this condition (Vesic, 2012). Consequently, it seemed that the space of attack relations satisfying this condition is rather narrow (note the similarity between direct undercut and direct defeat). However, the present paper shows that $\mathcal{R}_{ba}$ also belongs to (MC ↔ Ext), this indicating that the class of instantiations corresponding to the maxi-consistent operator is much larger.

The formal framework studied in this paper is that of classical propositional logic-based argumentation. The vast majority of ideas and considerations from the present paper hold for other instantiations of Dung's theory, for example in the setting studied by Modgil and Prakken (2013). In other words, the result obtained from those argumentation frameworks could also be compared with that obtained by an operator. After slightly adapting the definition of an operator, one can study the same questions: is there a link between the result obtained from an argumentation system and that obtained by an operator (from the same strict and defeasible rules)? Can argumentation help us find new operators? Are there argumentation systems returning a result that cannot be captured by an operator? Answering those questions will certainly be a part of our future work.

## Acknowledgments

The author would like to thank Leendert van der Torre for his assistance and advice. His useful comments helped to improve the paper significantly. The author also thanks the three reviewers for their helpful comments.

This paper is a revised and extended version of the conference paper: S. Vesic. Maxi-Consistent Operators in Argumentation. Proceedings of the 20th European Conference on Artificial Intelligence (ECAI'12), pages 810-815.

The major part of the work on this paper was carried out while the author was affiliated with the Computer Science and Communication Research Unit at the University of Luxembourg. First, the author started this work during the tenure of an ERCIM "Alain Bensoussan" Fellowship Programme, which is supported by the Marie Curie Co-funding of Regional, National and International Programmes (COFUND) of the European Commission. During this time, the author was also funded by the National Research Fund, Luxembourg. Second, while still at the University of Luxembourg, the author continued the work on this paper during a FNR AFR postdoc project which was supported by the National Research Fund, Luxembourg, and cofunded under the Marie Curie Actions of the European Commission (FP7-COFUND). Third, at the time when he was finishing the work on this paper, the author was a CRNS researcher affiliated with CRIL.